\PassOptionsToPackage{unicode=true}{hyperref}
\documentclass{easychair}

\usepackage{cite}

\usepackage{hyperref}
\usepackage{bussproofs}
\usepackage{tabularx}
\usepackage{booktabs}
\usepackage{multicol}
\usepackage[utf8]{inputenc}
\usepackage{stackengine}

\usepackage{subscript}
\newcolumntype{Y}{>{\centering\arraybackslash}X}
\usepackage{array}

\usepackage[ruled]{algorithm2e}
\usepackage{graphicx,grffile}
\usepackage{makecell}
\makeatletter
\def\maxwidth{\textwidth}
\def\maxheight{\ifdim\Gin@nat@height>\textheight\textheight\else\Gin@nat@height\fi}
\makeatother
\setkeys{Gin}{width=\maxwidth,height=\maxheight,keepaspectratio}
\usepackage{float}
\let\origfigure=\figure
\let\endorigfigure=\endfigure
\renewenvironment{figure}[1][]{%
\origfigure[!h]
}{%
\endorigfigure
}
\providecommand{\tightlist}{%
  \setlength{\itemsep}{0pt}\setlength{\parskip}{0pt}}
\usepackage{amsmath}
\usepackage{amssymb}
\usepackage{amsthm}

\newtheorem{definition}{Definition}
\newtheorem{example}{Example}

\providecommand{\citep}{\cite}
\DeclareRobustCommand{\href}[2]{#2\footnote{\url{#1}}}
\begin{document}

\title{On Improving the Backjump Level in PB Solvers}

\author{
	Romain Wallon
}

\institute{
    LIX (Laboratoire d'Informatique de l'X), École Polytechnique, X-Uber Chair\\
    \email{wallon@lix.polytechnique.fr}
}

\authorrunning{Romain Wallon}
\titlerunning{On Improving the Backjump Level in PB Solvers}

\maketitle

\begin{abstract}
Current PB solvers implement many techniques inspired by the CDCL
architecture of modern SAT solvers, so as to benefit from its practical
efficiency. However, they also need to deal with the fact that many of
the properties leveraged by this architecture are no longer true when
considering PB constraints. In this paper, we focus on one of these
properties, namely the optimality of the so-called first unique
implication point (1-UIP). While it is well known that learning the
first assertive clause produced during conflict analysis ensures to
perform the highest possible backjump in a SAT solver, we show that
there is no such guarantee in the presence of PB constraints. We also
introduce and evaluate different approaches designed to improve the
backjump level identified during conflict analysis by allowing to
continue the analysis after reaching the 1-UIP. Our experiments show
that sub-optimal backjumps are fairly common in PB solvers, even though
their impact on the solver is not clear.
\end{abstract}

\hypertarget{introduction}{%
\section{Introduction}\label{introduction}}

The CDCL architecture \citep{MarquesSilva1999} and the use of efficient
data structures and heuristics \citep{Moskewicz2001, Een2004} are at the
core of the practical efficiency of modern SAT solvers. Even though
these solvers can deal with very large benchmarks, containing millions
of variables and clauses, some relatively small problems (with only few
variables and clauses) remain completely out of their reach. This is
particularly true for problems that require the ability to ``count'',
such as those known as \emph{pigeonhole-principle formulae}, stating
that \(n\) pigeons cannot fit into \(n - 1\) holes. For such problems,
the resolution proof system used internally by SAT solvers is too weak:
it can only prove unsatisfiability with an exponential number of
derivation steps \citep{Haken1985}.

This has motivated the generalization of the CDCL architecture to handle
pseudo-Boolean (PB) problems \citep{Roussel2009}. Doing so, one can take
advantage of the strength of the \emph{cutting planes} proof system
\citep{Gomory1958}, which \emph{\(p\)-simulates} resolution: any
resolution proof can be translated into a cutting planes proof of
polynomial size w.r.t.~the size of the original proof. A common subset
of cutting planes is the one known as \emph{generalized resolution}
\citep{Hooker1988}. It is implemented in many PB solvers
\citep{Dixon2002, Chai2003, Sheini2006, LeBerre2010} to replace the use
of the resolution proof system during conflict analysis to learn new
constraints.

However, implementing the cutting planes proof system inside the CDCL
architecture also requires to deal with many broken invariants. For
instance, during conflict analysis, additional operations may be
required to ensure that the conflict is preserved after each derivation
step, and many different schemes have been proposed to this end
\citep{Elffers2018b, LeBerre2020}. The need for such operations makes
cutting planes-based PB solvers often slower in practice than
resolution-based SAT solvers, especially on CNF instances: generalized
resolution degenerates to resolution on such inputs.

In this paper, we consider another important difference between SAT
solvers and PB solvers, regarding the optimality of the \emph{first
unique implication point} (\(1\)-UIP). In SAT solvers, it is well known
that learning the first constraint that is assertive is optimal (i.e.,
gives the highest possible backjump level given the conflict)
\citep{Audemard2008}. It also provides a convenient and efficient
(syntactic) criterion for determining when the conflict analysis must be
stopped, that is, when only one literal in the derived clause is
assigned at the decision level at which the conflict occured. In PB
solvers, this is not the case anymore. First, determining whether a PB
constraint is assertive and at which level it is requires to perform
arithmetic operations on the coefficients, which may be costly because
of the use of arbitrary precision encodings. Second, learning the first
assertive constraint encountered during the analysis does not guarantee
to perform the highest possible backjump: we show that continuing the
analysis may allow to find a better (i.e., higher) backjump level.

In this paper, we investigate different approaches towards improving the
backjump levels computed by PB solvers. After having introduced the PB
solving framework in Section \ref{preliminaries}, we illustrate the
sub-optimality of the backjump levels computed by PB solvers in
Section~\ref{sub-optimal-backjumps-in-pb-solvers}. Based on these
examples, we introduce in Section
\ref{towards-an-improvement-of-the-backjump-level} different strategies
for improving the backjump levels in PB solvers by performing additional
cancellation steps after having derived an assertive constraint. In
particular, we present a criterion ensuring that the backjump level
never gets worse when the analysis continues. We also introduce
strategies allowing to detect when the analysis should stop. Finally, in
Section \ref{experimental-results}, we empirically evaluate these
approaches on the benchmarks of the PB competitions since their first
edition \citep{Manquinho2006}. We show that these new strategies allow
to improve the computed backjump levels, even though the impact of this
improvement on the performance of the solver is not clear.

\hypertarget{preliminaries}{%
\section{Preliminaries}\label{preliminaries}}

We consider a propositional setting defined on a finite set of
propositional variables \(\mathcal{V}\). A \emph{literal}~\(\ell\) is a
variable \(v \in \mathcal{V}\) or its negation \(\bar{v}\). Boolean
values are represented by the integers~\(1\)~(true) and \(0\)~(false),
so that \(\bar{v} = 1 - v\).

A \emph{pseudo-Boolean (PB) constraint} is an integral linear equation
or inequation over Boolean variables of the form
\(\sum_{i = 1}^{n} \alpha_i \ell_i \vartriangle \delta\), in which the
\emph{coefficients} (or \emph{weights}) \(\alpha_i\) and the
\emph{degree}~\(\delta\) are integers, \(\ell_i\) are literals and
\(\vartriangle \in \{ <, \leq, =, \geq, >\}\). Such a constraint can be
\emph{normalized} in linear time into a conjunction of constraints of
the form \(\sum_{i = 1}^{n} \alpha_i \ell_i \geq \delta\) in which the
coefficients and the degree are all non-negative integers. In the
following, we thus assume, without loss of generality, that all PB
constraints are normalized. A \emph{cardinality constraint} is a PB
constraint in which all coefficients are equal to~\(1\) and a
\emph{clause} is a cardinality constraint of degree \(1\). This
definition illustrates that PB~constraints are a generalization of
clauses, and that clausal reasoning is a special case of PB~reasoning.

PB solvers have thus been designed to extend the CDCL algorithm of
classical SAT solvers. In particular, when looking for a solution, PB
solvers need to \emph{assign} variables, either by \emph{making a
decision} or by \emph{propagating} a truth value. In the following, we
use the notation \(\ell (V@D)\) to represent that the literal \(\ell\)
has been assigned value \(V\) at decision level \(D\), and
\(\ell (?@?)\) to represent that \(\ell\) is unassigned. Let us remark
that, contrary to clauses, PB constraints may trigger propagations
multiple times, and even when some other literals are not assigned yet,
as illustrated below. Additionally, a PB constraint may also become
conflicting after having triggered a propagation at an earlier decision
level.

\begin{example}[{}]

\label{ex:propagatepb}

The PB constraint \(8a (?@?) + 2b (?@?) + c (?@?) + d (?@?) \geq 10\)
propagates the literal \(a\) at decision level \(0\). Later on, if \(d\)
becomes falsified, say at decision level \(3\), then the constraint
\(8a (1@0) + 2b (?@?) + c (?@?) + d (0@3) \geq 10\) propagates \(b\)
under the current partial assignment.

\end{example}

To detect whether a PB constraint is assertive (i.e., propagates some
literals), it is often convenient to compute the \emph{slack} of this
constraint \citep{Chai2003, Dixon2002}.

\begin{definition}[{Slack}]

\label{def:slack}

Let \(\chi\) be the PB constraint given by
\(\sum_{i=1}^n \alpha_i \ell_i \geq \delta\). The \emph{slack} of
\(\chi\) under the current partial assignment is the value
\(\sum_{i=1, \ell_i \neq 0}^n \alpha_i - \delta\).

\end{definition}

\begin{example}[{}]

\label{ex:slack}

The slack of the constraint
\(8a (1@0) + 2b (?@?) + c (?@?) + d (0@3)~\geq~10\) is
\(8 + 2 + 1 - 10 = 1\) at decision level \(3\).

\end{example}

Thanks to the slack of a constraint, it is possible to detect whether a
constraint is assertive or conflicting. More precisely, a constraint
propagates a literal \(\ell\) if the coefficient of \(\ell\) in the
constraint is strictly greater than the slack, and a constraint is
conflicting when its slack is negative.

\begin{example}[{Example \ref{ex:slack} cont'd}]

\label{ex:propagateslack}

At decision level \(3\), the slack of the constraint
\(8a (1@0) + 2b (?@?) + c (?@?) + d (0@3)~\geq~10\) is \(1\). This
constraint propagates thus \(b\) at decision level \(3\), since
\(2 > 1\).

\end{example}

When several assignments have been made by the solver, a \emph{conflict}
may occur (i.e., a constraint may be falsified). When this is the case,
a conflict analysis is performed by applying the \emph{cancellation}
rule between the conflicting constraint and the \emph{reason} for the
propagation of some of its literals, so as to derive a new constraint
(``LCM'' denotes the \emph{least common multiple}):

\begin{prooftree}

\AxiomC{$\alpha \ell + \sum_{i=1}^{n} \alpha_i \ell_i \geq \delta$}
\AxiomC{$\beta \bar{\ell} + \sum_{i=1}^{n} \beta_i \ell_i \geq \delta'$}
\AxiomC{$\mu\alpha = \nu\beta = \text{LCM}(\alpha, \beta)$}
\RightLabel{(canc.)}
\TrinaryInfC{$\sum_{i=1}^{n} (\mu\alpha_i + \nu\beta_i) \ell_i \geq \mu\delta + \nu\delta' - \mu\alpha$}

\end{prooftree}

Contrary to the resolution rule used in SAT solvers, this operation does
\emph{not} guarantee that the derived constraint will be conflicting. To
check that it will be the case, it is once again possible to use the
slack, which is particularly convenient as it is \emph{subadditive}: the
slack of the constraint obtained by applying the cancellation rule
between two constraints is at most equal to the sum of the slacks of
these constraints. This allows to estimate the slack of the constraint
that will be derived by providing an upper bound of its value: whenever
this upper bound is not negative, the constraint may be non-conflictual.
To preserve the conflict when this is the case, a possible approach is
to apply the \emph{weakening} and \emph{saturation} rules on the reason
until its slack becomes low enough to ensure the conflict to be
preserved (only literals that are not falsified may be weakened away)
\citep{Dixon2002}.

\begin{multicols}{2}
\begin{prooftree}
\AxiomC{$\alpha \ell + \sum_{i=1}^{n} \alpha_i \ell_i \geq \delta$}
\RightLabel{(weakening)}
\UnaryInfC{$\sum_{i=1}^{n} \alpha_i \ell_i \geq \delta - \alpha$}
\end{prooftree}

\begin{prooftree}
\AxiomC{$\sum_{i=1}^{n} \alpha_i \ell_i \geq \delta$}
\RightLabel{(saturation)}
\UnaryInfC{$\sum_{i=1}^{n} \min(\delta, \alpha_i) \ell_i \geq \delta$}
\end{prooftree}
\end{multicols}

The conflict analysis procedure of PB solvers consists of successive
applications of the cancellation rule (and of the weakening and
saturation rules when needed), following in reversed order the
assignments on the solver's trail (i.e., the assignment stack). We note
that current PB solvers do \emph{not} implement chronological
backtracking techniques as those presented in, e.g.,
\citep{Nadel2018, Moehle2019}, so that the assignments in the trail are
ordered by their decision level. The conflict analysis ends when the
first assertive constraint is derived, as in SAT solvers.

Since PB constraints may propagate different literals at different
decision levels (see Example~\ref{ex:propagatepb}), checking whether
such a constraint is assertive and at which level it is is harder than
when considering clauses or cardinality constraints. Indeed, in the case
of a PB constraint, one needs to compute, for each literal \(\ell\) of
the constraint, whether the constraint propagates a literal at the
decision level at which \(\ell\) had been assigned (based on the slack
at this decision level, for instance). Moreover, the PB constraint
produced by the conflict analysis procedure may be assertive at
different decision levels. In this case, the backjump must be performed
at the first decision level at which the constraint is assertive, as
current PB solvers always perform the propagations as soon as they
appear (once again, no chronological backtracking techniques are
implemented in such solvers).

\hypertarget{sub-optimal-backjumps-in-pb-solvers}{%
\section{Sub-Optimal Backjumps in PB
Solvers}\label{sub-optimal-backjumps-in-pb-solvers}}

The conflict analysis procedure described in the previous section takes
for granted the fact that learning a clause that is a \emph{first unique
implication point} (\emph{1-UIP}) provides the optimal backjump level.
While it is well known that, in SAT solvers, learning the first
assertive clause \emph{guarantees} to perform the highest possible
backjump \citep{Audemard2008}, this is not the case in PB solvers. In
particular, we exhibit in this section two examples of sub-optimal
backjump levels that may be computed when learning the first assertive
constraint.

\hypertarget{pigeonhole-principle-formulae}{%
\subsection{Pigeonhole-Principle
Formulae}\label{pigeonhole-principle-formulae}}

It is well known that, contrary to classical SAT solvers, PB solvers are
very efficient at solving pigeonhole-principle formulae. In particular,
while SAT solvers may need an exponential number of conflicts to prove
the unsatisfiability of such formulae, PB solvers based on cutting
planes can do so with a linear number of conflicts.

Let us solve such a problem with \(4\) pigeons and only \(3\) holes
following the approach of a PB solver. Consider the following encoding
of this problem, in which \(p_{i,j}\)
\((1 \leq i \leq 4, 1 \leq j \leq 3)\) represents that pigeon \(i\) is
put in hole \(j\):

\bigskip

\begin{multicols}{2}
$H_1 \equiv p_{1,1} + p_{2,1} + p_{3,1} + p_{4,1} \leq 1$

$H_2 \equiv p_{1,2} + p_{2,2} + p_{3,2} + p_{4,2} \leq 1$

$H_3 \equiv p_{1,3} + p_{2,3} + p_{3,3} + p_{4,3} \leq 1$

\columnbreak

$P_1 \equiv p_{1,1} + p_{1,2} + p_{1,3} \geq 1$

$P_2 \equiv p_{2,1} + p_{2,2} + p_{2,3} \geq 1$

$P_3 \equiv p_{3,1} + p_{3,2} + p_{3,3} \geq 1$

$P_4 \equiv p_{4,1} + p_{4,2} + p_{4,3} \geq 1$
\end{multicols}

\bigskip

The constraints on the left specify that ``each hole cannot contain more
than one pigeon'', while the constraints on the right specify that
``each pigeon must be put in a hole''. The normalized form of the
constraints above is given by:

\begin{multicols}{2}
$H_1 \equiv \bar p_{1,1} + \bar p_{2,1} + \bar p_{3,1} + \bar p_{4,1} \geq 3$

$H_2 \equiv \bar p_{1,2} + \bar p_{2,2} + \bar p_{3,2} + \bar p_{4,2} \geq 3$

$H_3 \equiv \bar p_{1,3} + \bar p_{2,3} + \bar p_{3,3} + \bar p_{4,3} \geq 3$

\columnbreak

$P_1 \equiv p_{1,1} + p_{1,2} + p_{1,3} \geq 1$

$P_2 \equiv p_{2,1} + p_{2,2} + p_{2,3} \geq 1$

$P_3 \equiv p_{3,1} + p_{3,2} + p_{3,3} \geq 1$

$P_4 \equiv p_{4,1} + p_{4,2} + p_{4,3} \geq 1$
\end{multicols}

\bigskip

Let us now try to assign some variables, until we get a conflict. We
first assign \(p_{1,1}\) to \(0\) at decision level \(1\) and
\(p_{1,2}\) to \(0\) at decision level \(2\). At this point, we have
that:

\begin{itemize}
\tightlist
\item
  \(P_1\) propagates \(p_{1,3}\) to \(1\), and
\item
  \(H_3\) propagates then \(p_{2,3}\), \(p_{3,3}\) and \(p_{4,3}\) to
  \(0\).
\end{itemize}

If we now assign \(p_{2,1}\) to \(0\), we have that:

\begin{itemize}
\tightlist
\item
  \(P_2\) propagates \(p_{2,2}\) to \(1\),
\item
  \(H_2\) propagates then \(p_{3,2}\) and \(p_{4,2}\) to \(0\),
\item
  \(P_3\) and \(P_4\) now propagate \(p_{3,1}\) and \(p_{4,1}\) to
  \(1\), respectively, and
\item
  \(H_1 \equiv \bar p_{1,1} (1@1) + \bar p_{2,1} (1@3) + \bar p_{3,1} (0@3) + \bar p_{4,1} (0@3) \geq 3\)
  is conflicting.
\end{itemize}

The conflict is thus analyzed by applying successively the cancellation
rule between the conflicting constraint \(H_1\) and the reasons for
\(p_{4,1}\) and \(p_{3,1}\) being propagated to \(1\), i.e., \(P_4\) and
\(P_3\), respectively.

\begin{prooftree}

\AxiomC{$H_1$}
\AxiomC{$P_4$}
\BinaryInfC{$\bar p_{1,1} (1@1) + \bar p_{2,1} (1@3) + \bar p_{3,1} (0@3) + p_{4,2} (0@3) + p_{4,3} (0@2) \geq 3$}

\AxiomC{$P_3$}
\BinaryInfC{$\bar p_{1,1} (1@1) + \bar p_{2,1} (1@3) + p_{3,2} (0@3) + p_{3,3} (0@2) + p_{4,2} (0@3) + p_{4,3} (0@2) \geq 3$}

\end{prooftree}

The constraint is still not assertive, so we apply the cancellation rule
between the inferred constraint and the reason for both \(p_{3,2}\) and
\(p_{4,2}\) being propagated to \(0\), i.e., \(H_2\).

\begin{prooftree}

\AxiomC{$\bar p_{1,1} (1@1) + \bar p_{2,1} (1@3) + p_{3,2} (0@3) + p_{3,3} (0@2) + p_{4,2} (0@3) + p_{4,3} (0@2) \geq 3$}
\AxiomC{$H_2$}
\BinaryInfC{$\bar p_{1,1} (1@1) + \bar p_{1,2} (1@2) + \bar p_{2,1} (1@3) + \bar p_{2,2} (0@3) + p_{3,3} (0@2) + p_{4,3} (0@2) \geq 4$}

\end{prooftree}

Observe that this constraint propagates \(p_{2,1}\) and \(p_{2,2}\) to
\(1\) at decision level \(2\). This constraint is thus assertive, so the
solver learns it, and backjumps to decision level \(2\), before
continuing to explore the search space. But can we do better than this?

\bigskip

Suppose we continue the analysis, and apply the cancellation rule
between the assertive constraint above and the reason for \(p_{2,2}\)
being assigned to \(1\), i.e., \(P_2\).

\begin{prooftree}

\AxiomC{$\bar p_{1,1} (1@1) + \bar p_{1,2} (1@2) + \bar p_{2,1} (1@3) + \bar p_{2,2} (0@3) + p_{3,3} (0@2) + p_{4,3} (0@2) \geq 4$}
\AxiomC{$P_2$}
\BinaryInfC{$\bar p_{1,1} (1@1) + \bar p_{1,2} (1@2) + p_{2,3} (0@2) + p_{3,3} (0@2) + p_{4,3} (0@2) \geq 3$}

\end{prooftree}

We note that there is no more literal assigned at decision level \(3\)
in the derived constraint, but it is \emph{conflicting} at decision
level \(2\). This is not so surprising, as finding a conflict at a
higher decision level may also happen when performing a ``regular''
conflict analysis in a PB solver. Let us thus continue to analyze this
conflict. The next cancellation to perform is with the reason for
\(p_{2,3}\), \(p_{3,3}\) and \(p_{4,3}\) being assigned to \(0\), i.e.,
\(H_3\).

\begin{prooftree}

\AxiomC{$\bar p_{1,1} (1@1) + \bar p_{1,2} (1@2) + p_{2,3} (0@2) + p_{3,3} (0@2) + p_{4,3} (0@2) \geq 3$}
\AxiomC{$H_3$}
\BinaryInfC{$\bar p_{1,1} (1@1) + \bar p_{1,2} (1@2) + \bar p_{1,3} (0@2) \geq 3$}

\end{prooftree}

This constraint is assertive at decision level \(0\), and propagates to
\(0\) the variables \(p_{1,1}\), \(p_{1,2}\) and \(p_{1,3}\). We can
still perform a last cancellation step, between the constraint above and
the reason for \(p_{1,3}\) being propagated to \(1\), i.e., \(P_1\).

\begin{prooftree}

\AxiomC{$\bar p_{1,1} (1@1) + \bar p_{1,2} (1@2) + \bar p_{1,3} (0@2) \geq 3$}
\AxiomC{$P_1$}
\BinaryInfC{$0 \geq 1$}

\end{prooftree}

We have thus been able to derive \(0 \geq 1\) (i.e., \(\bot\)) in a
\emph{single conflict analysis} for this pigeonhole instance. Actually,
it is easy to see that, for any pigeonhole instance with \(n\) pigeons
to put in \(n - 1\) holes, it is always possible to prove
unsatisfiability with a single conflict if we do not stop the analysis
when the first assertive constraint is derived. This also shows that
learning the first assertive constraint is \emph{not optimal in
general}. However, continuing conflict analysis does not guarantee to
always improve the backjump level, as shown in the next example.

\hypertarget{choosing-the-right-constraints}{%
\subsection{\texorpdfstring{Choosing the \emph{Right}
Constraints}{Choosing the Right Constraints}}\label{choosing-the-right-constraints}}

As already mentioned, it is well known that it is not possible to find a
higher backjump level than that provided by the 1-UIP in SAT solvers. In
the following example, we show that the backjump level may also increase
when performing additional cancelletion steps with PB constraints.

Suppose that the solver is run on a given instance, and that, during the
search, the constraint
\(\chi \equiv 4a(0@10) + 4b(1@30) + 3c(0@20) + 3d(0@30) + 2e(1@30) + f(0@40) + g(0@40) + z(0@40) \geq 8\)
becomes conflicting. Suppose that the reason for \(f\) and \(g\) being
propagated to \(0\) is
\(\rho_1~\equiv~3i(0@20) + 3j(0@40) + 2\bar{f}(1@40) + 2\bar{g}(1@40) + h(1@40) \geq 5\).
The cancellation rule is applied between these two constraints.

\begin{prooftree}

\AxiomC{$\chi$}
\AxiomC{$\rho_1$}
\BinaryInfC{\stackanchor{$8a(0@10) + 8b(1@30) + 6c(0@20) + 6d(0@30) + 4e(1@30)$}{$+ 2z(0@40) + 3i(0@20) + 3j(0@40) + h(1@40) \geq 17$}}

\end{prooftree}

Let \(\chi'\) be the constraint derived above. \(\chi'\) is assertive,
and propagates \(b\) at decision level~\(20\). This constraint is thus
learned, and a backjump is performed. Once again, let us try to find a
higher backjump level. Suppose that the reason for \(j\) is
\(\rho_2 \equiv 6\bar{c}(1@20) + 6\bar{d}(1@30) + 3\bar{j}(1@40) + 3k(0@40) + 3l(0@30) \geq 15\),
and let us apply the cancellation rule between this constraint
and~\(\chi'\).

\begin{prooftree}

\AxiomC{$\chi'$}
\AxiomC{$\rho_2$}
\BinaryInfC{\stackanchor{$8a(0@10) + 8b(1@30) + 4e(1@30) + 2z(0@40)$}{$+ 3i(0@20) + 3k(0@40) + 3l(0@30) + h(1@40) \geq 17$}}

\end{prooftree}

The constraint obtained here is also assertive and propagates now \(b\)
at decision level \(10\). We thus managed to improve the backjump level.
However, suppose that, instead of applying the cancellation rule with
the reason for \(j\) above, we had applied it with the reason for \(z\)
given by
\(\rho_3 \equiv 10w(1@25) + 10x(0@25) + y(0@40) + \bar{z}(1@40) \geq 11\).

\begin{prooftree}

\AxiomC{$\chi'$}
\AxiomC{$\rho_3$}
\BinaryInfC{\stackanchor{$20w(1@25) + 20x(0@25) + 8a(0@10) + 8b(1@30) + 6c(0@20) + 6d(0@30)$}{$+ 4e(1@30) + 3i(0@20) + 3j(0@40) + 2y(0@40) + h(1@40) \geq 37$}}

\end{prooftree}

Even though the constraint is still assertive, it propagates now \(b\)
at decision level \(25\). So, continuing conflict analysis may also
worsen the backjump level, as it is the case for clauses. As shown in
this example, it may even depend on the order of the propagations on the
trail. To make sure that continuing the analysis never worsens the
backjump level, we thus need to design appropriate criteria for deciding
\emph{when} to continue.

\hypertarget{towards-an-improvement-of-the-backjump-level}{%
\section{Towards an Improvement of the Backjump
Level}\label{towards-an-improvement-of-the-backjump-level}}

In this section, we introduce an approach for improving the backjump
level computed by PB solvers by allowing to continue the analysis after
an assertive constraint has been derived. To this end, we extend the
classical conflict analysis procedure of PB solvers by considering two
new criteria that are used after the first assertive constraint has been
produced to decide whether additional cancellations should be performed,
and on which constraints. These criteria are designed to make sure that
an assertive constraint will indeed be derived at the end of the
analysis, while optimizing the height of the computed backjump level.

\hypertarget{non-worsening-backjump-level}{%
\subsection{Non-Worsening Backjump
Level}\label{non-worsening-backjump-level}}

As observed in the previous section, continuing the analysis does not
guarantee to improve the computed backjump level. In the worst case, the
assertion may even be lost, which would break the CDCL algorithm. We
thus need to introduce a new invariant for the conflict analysis
procedure: if the conflicting constraint is assertive at decision level
\(b\), the constraint that will be obtained after applying the
cancellation rule between this constraint and a reason must be either
assertive at decision level \(b\) or conflicting at decision level
\(b\). It is easy to see that such an invariant guarantees that an
assertive constraint at decision level \(b\) or higher will be
eventually derived. Let us now describe how to maintain it.

Suppose that the conflicting constraint is assertive at decision level
\(b\), and that we want to perform a cancellation with a reason. The
resulting constraint will be assertive at decision level~\(b\) if it
propagates (at least) one of its literals. In particular, (at least) one
of its literals must have a coefficient that is strictly greater than
the slack of this constraint \emph{computed at decision level \(b\)}. We
note that, in this particular case, it is preferable to compute the
\emph{exact} slack (rather than using the subadditive property), to
limit the number of false negative. Similarly, the slack may be used to
check whether the constraint will be conflicting at decision level
\(b\).

When the value of the slack guarantees that the constraint will remain
assertive or conflicting, the cancellation rule can be applied safely.
Otherwise, one can simply ignore the reason and go on to the next
propagation on the trail (as a ``regular'' conflict analysis would have
ignored it anyway). Another solution is to use an approach similar to
that used to preserve conflicts during a ``regular'' conflict analysis.
More precisely, one can try to weaken away some of the literals of the
reason until the constraint obtained after applying the cancellation is
assertive or conflicting at decision level \(b\).

\begin{example}[{}]

Suppose that, during conflict analysis, the constraint
\(4\bar{a} (0@5) + 4\bar{b} (0@3) + 4\bar{c} (0@4) + 4\bar{d} (0@3) + e (0@1) + f (0@2) + \bar{g} (1@4) + \bar{h} (1@5) \geq 4\)
is derived. This constraint is assertive at decision level \(4\), and
propagates \(a\) to \(0\) at this decision level. Suppose now that the
reason for \(a\) being propagated to \(1\) is the constraint
\(2a (1@5) + 2b (1@3) + 2c (1a4) + 2g (0@4) + 2h (0@5) \geq 5\). If we
resolve this reason with the assertive constraint above, the derived
constraint is a tautology, and is thus not assertive at any decision
level. However, if the literals \(b\) and \(c\) are weakened away from
the reason to get the clause \(a (1@5) + g (0@4) + h (0@5) \geq 1\)
(which still propagates \(a\) at decision level \(5\)), the constraint
obtained after applying the cancellation with the assertive constraint
above is
\(2\bar{b} (0@3) + 2\bar{c} (0@4) + 2\bar{d} (0@3) + 2g (0@4) + 2h (0@5) + e (0@1) + f (0@2) \geq 2\)
which is still assertive at decision level \(4\), and propagates now
\(h\) to \(1\) at this decision level.

\end{example}

Note that, contrary to the case in which the weakening and saturation
rules are applied to preserve the conflict, there is no guarantee that
applying these rules will eventually preserve the assertivity or the
conflict at decision level \(b\). In particular, the successive
weakening operations may produce a tautology from the reason, and the
cancellation step must be ignored. We also note that this is compatible
with the 1-UIP scheme in SAT solvers, as weakening a literal from a
clause always produces a tautology.

Remark that, when a new assertive constraint is derived, the approach
above guarantees that it propagates a literal at decision level \(b\).
However, this literal may be propagated at a decision level \(b'\) that
is \emph{higher} than \(b\). This is precisely what allows to
\emph{improve} the backjump level. For the subsequent application of the
cancellation rule, the decision level \(b'\) must thus be used instead
of \(b\) when ensuring to maintain the invariant.

\hypertarget{ending-conflict-analysis}{%
\subsection{Ending Conflict Analysis}\label{ending-conflict-analysis}}

Since deriving an assertive constraint is no longer a sufficient stop
condition for the conflict analysis, we need a new criterion for
determining when to stop it.

We remark that, if the assertive constraint we have derived is assertive
at decision level \(b\), we should not apply the cancellation rule
between this constraint and reasons for literals propagated at decision
level \(b\) or higher. While doing so may still allow to find a better
backjump level, there is no guarantee that it will be the case, which
may lead to practical issues. Indeed, as assignments are undone at each
cancellation step (even those that are ignored), it would be needed to
\emph{redo} all of them if the backjump level remains the same, and the
backjump would become a \emph{forejump}, which seems unreasonable. So, a
naive criterion is to stop the analysis when the highest decision level
on the trail is the latest computed backjump level.

Of course, this does not apply to the particular case in which the
derived constraint has become conflicting at decision level \(b\). In
this case, a ``regular'' conflict analysis must start again with the new
conflicting constraint.

Let us note that the criterion above may be optimized in different ways.
For instance, if at some point the derived assertive constraint
propagates a literal at decision level \(0\), it will not be possible to
find a higher backjump level, so the analysis can stop (even though one
could still derive \(\bot\), as in the pigeonhole-principle example
presented in the previous section). Similarly, if the constraint is
conflicting at decision level \(0\), the analysis can stop early, as the
problem is unsatisfiable in this case. We could also decide to stop the
analysis when we consider the backjump level to be ``high enough'',
based on (parameterizable) numerical criteria.

We remark that the stopping criteria presented in this section are
designed towards improving the backjump level computed by the conflict
analysis procedure. However, different criteria may also be considered
for determining when to stop the analysis, such as the quality of the
derived constraint (e.g., with different quality measures as introduced
in \citep{LeBerre2021}), or its propensity to generate short proofs
(e.g., by ensuring to derive constraints having a low slack).

\hypertarget{experimental-results}{%
\section{Experimental Results}\label{experimental-results}}

In this section, we present an empirical evaluation of the approach
described in Section \ref{towards-an-improvement-of-the-backjump-level}.
All experiments presented in this section have been run on a cluster
equiped with quadcore bi-processors Intel XEON E5-5637 v4 (3.5 GHz) and
128 GB of memory. The time limit was set to 1200 seconds and the memory
limit to 32~GB. The whole set of decision benchmarks containing
``small'' integers used in the PB competitions since the first edition
\citep{Manquinho2006} was considered as input.

\hypertarget{implementation-details}{%
\subsection{Implementation Details}\label{implementation-details}}

To evaluate our approach, we implemented it in
\emph{Sat4j-CuttingPlanes} \citep{LeBerre2010}. This implementation,
which is available in a
\href{https://gitlab.ow2.org/romain_wallon/sat4j/-/tree/assertion-level}{dedicated
repository}, provides different variants of the two criteria presented
in the previous section. They are described below.

In order to preserve the invariant that the derived constraint remains
either assertive at a decision level \(b\) or conflicting at this
decision level, we experimented the following variants:

\begin{itemize}
\tightlist
\item
  \texttt{never-weaken}: if the invariant is not preserved, no weakening
  is applied and the cancellation step is ignored.
\item
  \texttt{weaken-any}: if the invariant is not preserved, a literal that
  is not falsified at decision level~\(b\) is weakened iteratively until
  the invariant is restored or the reason becomes a tautology (in which
  case the cancellation step is ignored).
\item
  \texttt{weaken-ordered}: similar to \texttt{weaken-any}, but weakened
  literals are preferably those unassigned or, assigned at the lowest
  decision level (in this order).
\end{itemize}

We also implemented different strategies for detecting when to stop the
conflict analysis after having derived an assertive constraint. Suppose
that the first literal propagated by this constraint is propagated at
decision level \(b\). Our strategies are:

\begin{itemize}
\tightlist
\item
  \texttt{until-bjlevel}: the conflict analysis stops when the highest
  decision level on the trail is~\(b\) (i.e., when all assignments made
  at a decision level higher than \(b\) have been undone).
\item
  \texttt{until-toplevel}: the conflict analysis stops when the highest
  decision level on the trail is \(b\), or when \(b\) is \(0\).
\item
  \texttt{until-highlevel}: the conflict analysis stops when the highest
  decision level on the trail is \(b\), or when \(b\) is \(10\%\) of the
  highest decision level on the trail, so as to avoid making too many
  additional (and costly) cancellation steps.
\end{itemize}

We evaluated all these strategies, and the complete analysis of their
results may be retrieved in a
\href{https://gitlab.com/pb-backjump-level/pos21-experiments}{dedicated
repository}. As there is no clear difference between them, this section
only reports the results of the combination of \texttt{weaken-any} and
\texttt{until-bjlevel}.

\hypertarget{effectiveness-of-our-approach}{%
\subsection{Effectiveness of our
Approach}\label{effectiveness-of-our-approach}}

\begin{figure}
\centering
\includegraphics[width=\textwidth,height=0.95\textheight]{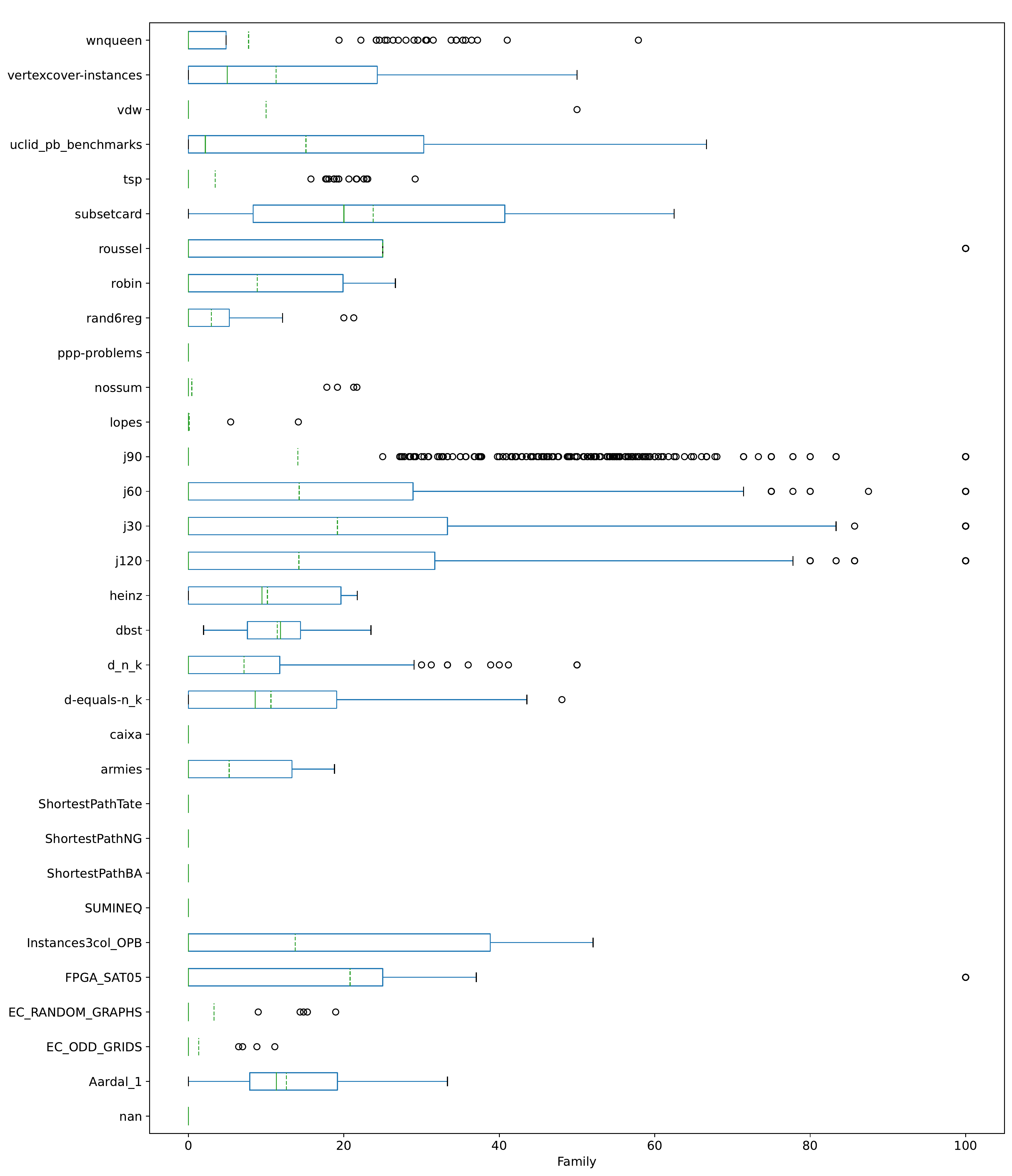}
\caption{Boxplots of the percentage of improved backjumps per family.
Each box displays the quartiles with the vertical bars and the estimated
minimum and maximum with the horizontal bars, computed from the
percentage of improved backjumps per instance in each family. Points
represent outliers, which are instances for which the percentage of
improved backjumps is above the estimated maximum.
\label{fig:suboptimal}}
\end{figure}

Let us first analyze to what extent continuing conflict analysis after
having produced an assertive constraint allows to find higher backjump
levels. To this end, we consider the percentage of conflict analyses for
which our approach allowed to improve the backjump level. The results
are given in Figure \ref{fig:suboptimal}.

The boxplots on this figure show that our approach indeed allows to
improve the backjump levels initially computed with the first assertive
constraint, and that this is the case for many families of benchmarks.
Moreover, we can see that for some families, this happens quite often in
practice. For instance, for the \texttt{subsetcard} family, more than
\(20\%\) of the computed backjump levels are sub-optimal in average (and
the median is also near \(20\%\)).

Let us remark that, as constraints learned by the solver differ when
using our approach, the boxplots here does \emph{not} represent the
percentage of sub-optimal backjump levels performed by the solver when
it applies a ``regular'' conflict analysis, but those that we were able
to \emph{improve} with our approach.

\hypertarget{impact-of-the-improved-backjump-levels}{%
\subsection{Impact of the Improved Backjump
Levels}\label{impact-of-the-improved-backjump-levels}}

\begin{figure}
\centering
\includegraphics[width=\textwidth,height=0.3\textheight]{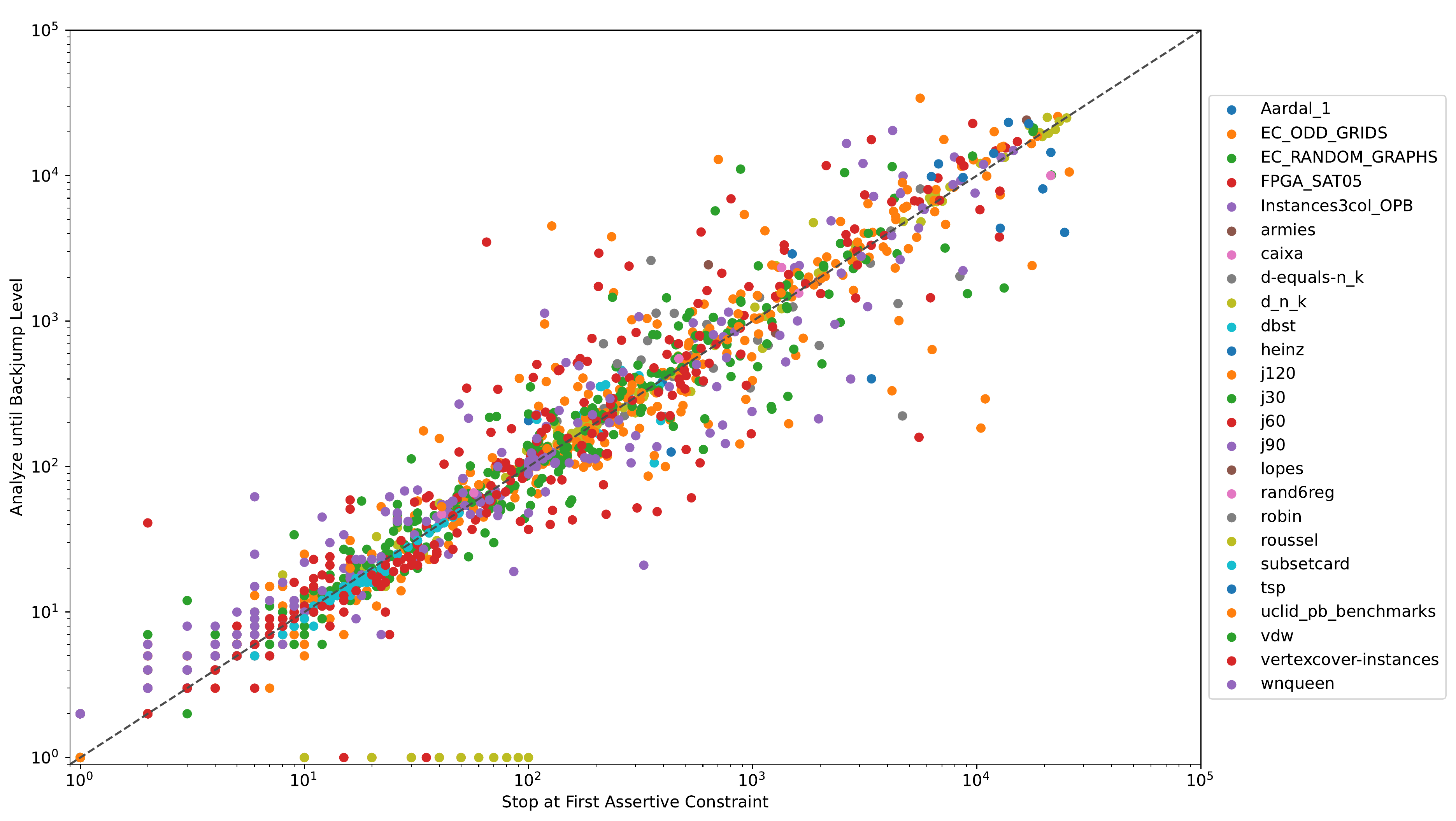}
\caption{Scatter plot of the number of conflicts. Only instances that
are solved by both approaches are reported, as the proof must be
complete for a fair comparison.\label{fig:conflicts}}
\end{figure}

\begin{figure}
\centering
\includegraphics[width=\textwidth,height=0.3\textheight]{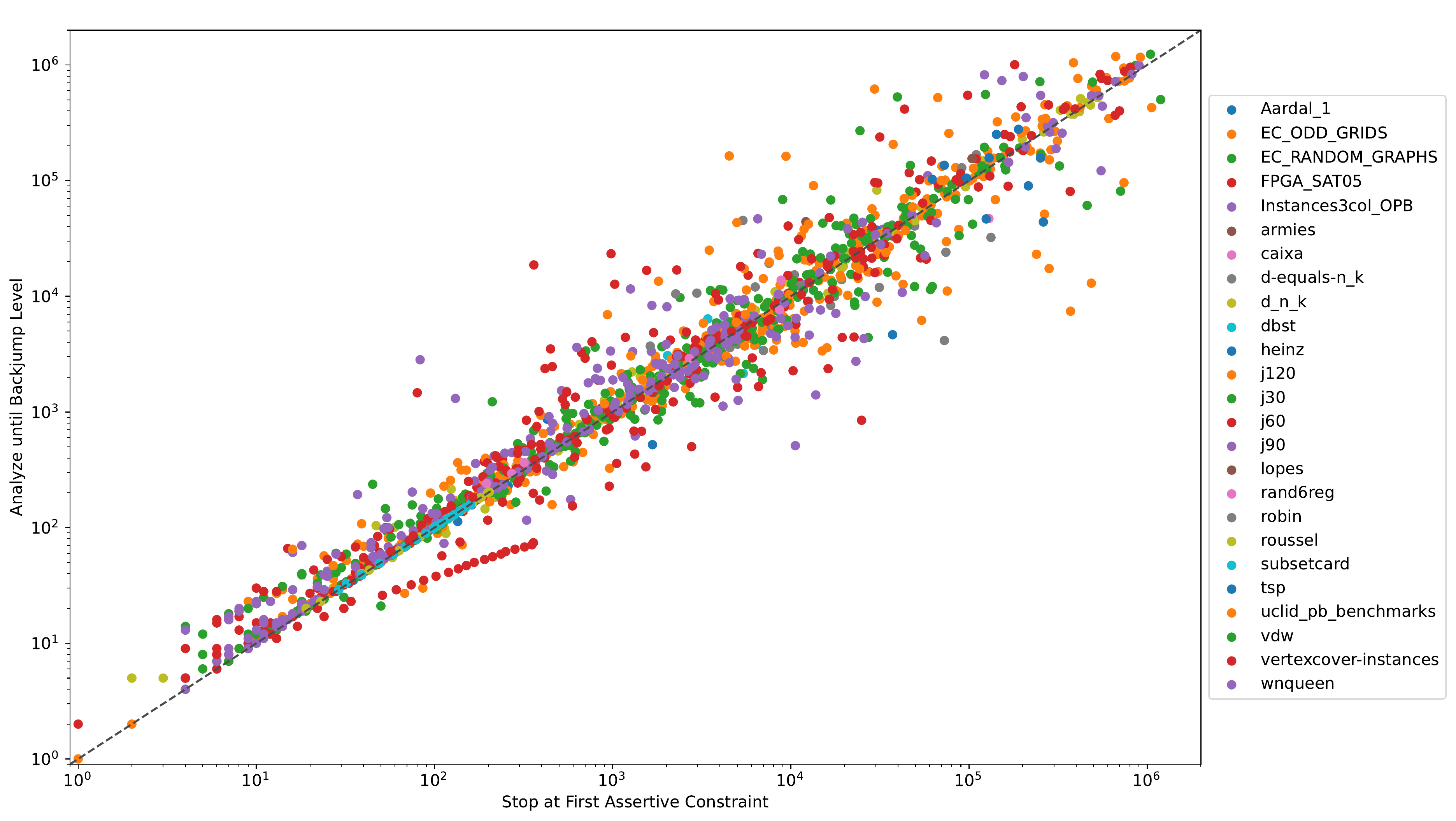}
\caption{Scatter plot of the number of cancellations. Only instances
that are solved by both approaches are reported, as the proof must be
complete for a fair comparison.\label{fig:cancellations}}
\end{figure}

In order to evaluate the impact of the improved backjump levels, we
first consider the size of the proofs built by the solver, as the
approach we presented is penalized by the additional operations it
requires (especially, the computation of the \emph{exact} slack of the
constraints). The results are given in Figures \ref{fig:conflicts} and
\ref{fig:cancellations}.

Figure \ref{fig:conflicts} shows that, as expected, pigeonhole instances
(from the \texttt{roussel} family) are always solved with a single
conflict. Quite interestingly, the number of cancellations remains the
same on these instances: in fact the same proof is built by a
``regular'' conflict analysis, but several conflicts are required to
find it.

In Figure \ref{fig:cancellations}, we can see that the proof built by
the solver on instances of the \texttt{vertexcover} family is
exponentially shorter when the conflict analysis continues. We looked
closer at the behavior of the solver on these instances, and what we
observed is that the first conflict that is analyzed occurs on the
constraint \(\sum_{i=1}^n 2 \bar x_i \geq n\). During the analysis of
this conflict, the cancellation rule is successively applied between
this constraint and \emph{all} clauses of the form \(x_1 + x_j \geq 1\)
(\(2 \leq j \leq n\)), producing \(x_1 \geq 1\). As noticed in
\citep{LeBerre2020a}, the constraint that is learned when the conflict
analysis stops at the first assertive constraint contains a number of
irrelevant literals that cause the proof to be exponentially larger.
When the analysis continues, all these irrelevants literals get
cancelled out, thus allowing to find the shorter proof found when
removing explicitly irrelevant literals.

Regarding the other families, there is no clear difference between
continuing the analysis or stopping it at the first assertive
constraint. For some families, continuing the analysis makes almost no
difference. For some other families, there exist instances for which the
proof is shorter when continuing the analysis, and others for which the
proof is shorter when stopping at the first assertive constraint. This
suggests that we lack an additional heuristic for determining when the
analysis should continue, or that our criterion for determining when to
stop the analysis is not good enough.

\bigskip

\begin{figure}
\centering
\includegraphics[width=\textwidth,height=0.45\textheight]{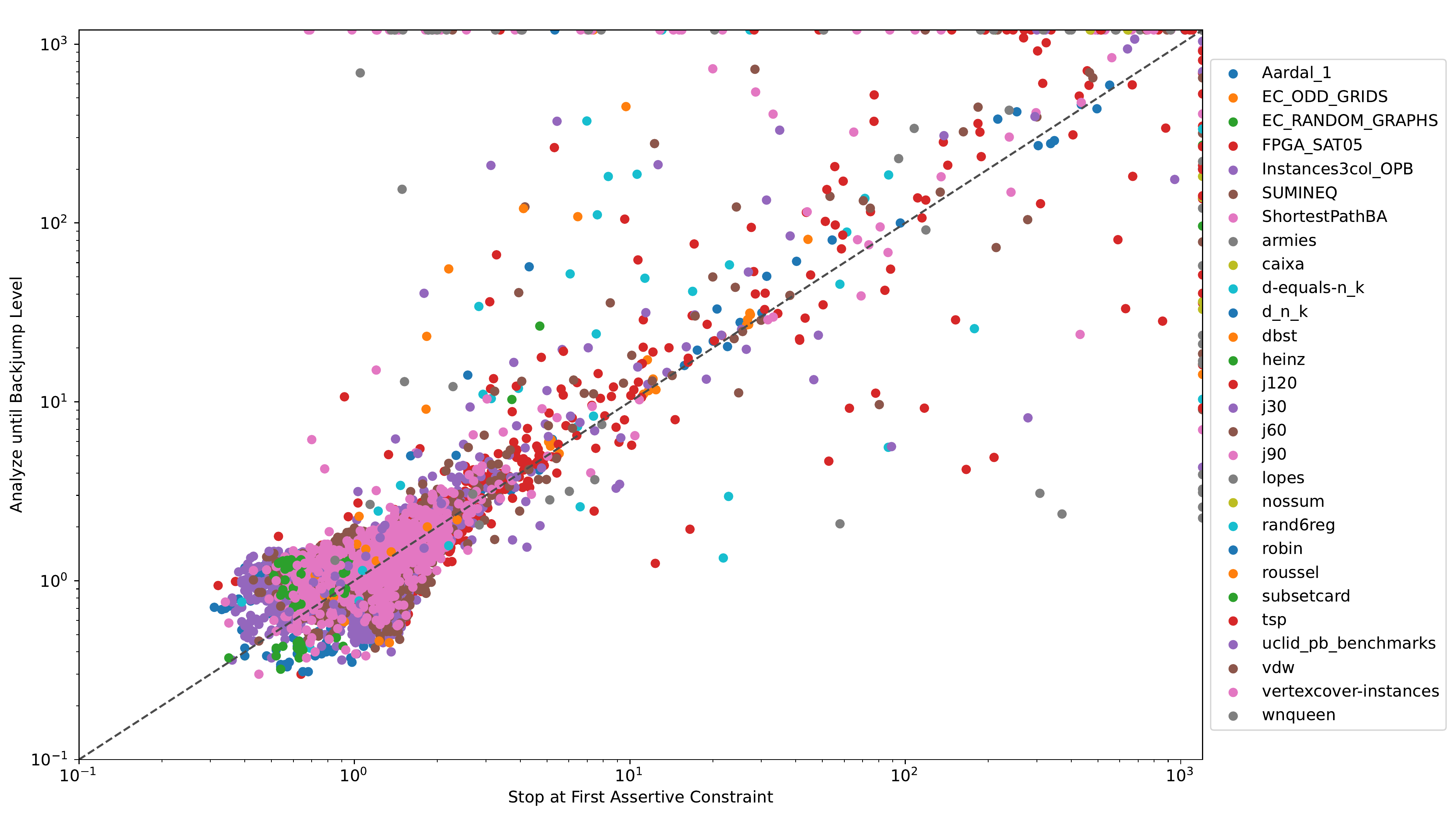}
\caption{Scatter plot of the runtime.\label{fig:runtime}}
\end{figure}

Let us now compare the runtime of our approach with that of the solver
when it runs a ``regular'' conflict analysis. The results are given in
Figure \ref{fig:runtime}. Clearly, our approach remains costly in many
cases, especially because checking that our invariant is preserved
requires to perform a number of arithmetic operations, whose cost is
increased by the use of arbitrary precision encodings. On the scatter
plot, we can observe that only few instances are solved faster with our
approach. These instances are those for which the size of the proof is
reduced enough to counter-balance the added cost of the additional
operations.

\clearpage

\hypertarget{conclusion}{%
\section{Conclusion}\label{conclusion}}

In this paper, we showed that, contrary to classical SAT solvers, the
backjump level computed by PB solvers when learning the first assertive
constraint derived during conflict analysis is not optimal in general.
More precisely, if the conflict analysis continues after reaching such a
constraint, it is possible to decrease the backjump level, and thus to
backjump higher in the search tree. To make it possible, we presented
and evaluated different strategies designed to preserve the invariants
of the solver and to make sure to never worsen the computed backjump
level while performing additional cancellation steps. While doing so
indeed allows to improve the backjump levels computed during conflict
analysis, the impact of this approach on the performance of the solver
is unclear.

A perspective for future work is to find heuristics to decide
\emph{when} the conflict analysis should be continued after having
derived the first assertive constraint and when it should not, for
instance based on the quality of the derived constraints or on the
(expected) size of the proof. Improving our different strategies for
choosing on \emph{which} constraints the additional cancellation steps
should be applied is also an important perspective, especially to make
them more efficient. It would also be interesting to investigate how to
combine these strategies with chronological backtracking techniques that
have recently been investigated for SAT solvers, but not exploited yet
in PB solvers. Another avenue to explore is to find ways of continuing
the analysis \emph{asynchronously}, so as to allow the solver to learn
the first assertive constraint and backjump as usual, while the learned
constraint is refined to trigger a better backjump, only if it is worth
doing so.

\section*{Acknowledgements}
The author is grateful to the anonymous reviewers for their numerous
comments, that greatly helped to improve the paper. The author is also
grateful to the CRIL laboratory (CNRS UMR 8188, Université d'Artois) for
having supplied the computational resources on which the experiments
presented in this paper have been run. This publication was supported by
the Chair ``Integrated Urban Mobility'', backed by L'X -- École
Polytechnique and La Fondation de l'École Polytechnique and sponsored by
Uber. The Partners of the Chair shall not under any circumstances accept
any liability for the content of this publication, for which the author
shall be solely liable.

\bibliographystyle{plain}
\bibliography{bibliography/bibliography.bib}

\begin{thebibliography}{10}

\bibitem{Audemard2008}
Gilles Audemard, Lucas Bordeaux, Youssef Hamadi, Said Jabbour, and Lakhdar
  Sais.
\newblock {A Generalized Framework for Conflict Analysis}.
\newblock Technical report, February 2008.

\bibitem{Chai2003}
Donald Chai and Andreas Kuehlmann.
\newblock {A fast pseudo-boolean constraint solver}.
\newblock In {\em Proceedings of the 40th Design Automation Conference, {DAC}
  2003, Anaheim, CA, USA, June 2-6, 2003}, pages 830--835. {ACM}, 2003.

\bibitem{Dixon2002}
Heidi~E. Dixon and Matthew~L. Ginsberg.
\newblock Inference methods for a pseudo-boolean satisfiability solver.
\newblock In {\em AAAI'02}, pages 635--640, 2002.

\bibitem{Een2004}
Niklas E{\'e}n and Niklas S{\"o}rensson.
\newblock An extensible sat-solver.
\newblock In {\em Theory and Applications of Satisfiability Testing}, pages
  502--518, 2004.

\bibitem{Elffers2018b}
Jan Elffers and Jakob Nordstr\"{o}m.
\newblock Divide and conquer: Towards faster pseudo-boolean solving.
\newblock In {\em Proceedings of the Twenty-Seventh International Joint
  Conference on Artificial Intelligence, {IJCAI-18}}, pages 1291--1299, 2018.

\bibitem{Gomory1958}
Ralph~E. Gomory.
\newblock Outline of an algorithm for integer solutions to linear programs.
\newblock {\em Bulletin of the American Mathematical Society}, pages 275--278,
  1958.

\bibitem{Haken1985}
Armin Haken.
\newblock {The intractability of resolution}.
\newblock {\em {Theoretical Computer Science}}, 39:297 -- 308, 1985.
\newblock {Third Conference on Foundations of Software Technology and
  Theoretical Computer Science}.

\bibitem{Hooker1988}
J.~N. Hooker.
\newblock Generalized resolution and cutting planes.
\newblock {\em Annals of Operations Research}, 12(1):217--239, 1988.

\bibitem{LeBerre2020a}
Daniel {Le Berre}, Pierre Marquis, Stefan Mengel, and Romain Wallon.
\newblock On irrelevant literals in pseudo-boolean constraint learning.
\newblock In Christian Bessiere, editor, {\em Proceedings of the Twenty-Ninth
  International Joint Conference on Artificial Intelligence, {IJCAI} 2020},
  pages 1148--1154. ijcai.org, 2020.

\bibitem{LeBerre2020}
Daniel {Le Berre}, Pierre Marquis, and Romain Wallon.
\newblock On weakening strategies for {PB} solvers.
\newblock In Luca Pulina and Martina Seidl, editors, {\em Theory and
  Applications of Satisfiability Testing - {SAT} 2020 - 23rd International
  Conference, Alghero, Italy, July 3-10, 2020, Proceedings}, volume 12178 of
  {\em Lecture Notes in Computer Science}, pages 322--331. Springer, 2020.

\bibitem{LeBerre2010}
Daniel Le~Berre and Anne Parrain.
\newblock {The SAT4J library, Release 2.2, System Description}.
\newblock {\em {Journal on Satisfiability, Boolean Modeling and Computation}},
  7:59--64, 2010.

\bibitem{LeBerre2021}
Daniel Le~Berre and Romain Wallon.
\newblock On dedicated cdcl strategies for pb solvers.
\newblock In {\em Theory and Applications of Satisfiability Testing - {SAT}
  2021 - 24th International Conference, Proceedings}, page to appear, 2021.

\bibitem{Manquinho2006}
Vasco Manquinho and Olivier Roussel.
\newblock The first evaluation of pseudo-boolean solvers (pb'05).
\newblock {\em JSAT}, pages 103--143, 2006.

\bibitem{MarquesSilva1999}
Joao Marques-Silva and Karem~A. Sakallah.
\newblock Grasp: A search algorithm for propositional satisfiability.
\newblock {\em IEEE Trans. Computers}, pages 220--227, 1999.

\bibitem{Moehle2019}
Sibylle M{\"{o}}hle and Armin Biere.
\newblock Backing backtracking.
\newblock In Mikol{\'{a}}s Janota and In{\^{e}}s Lynce, editors, {\em Theory
  and Applications of Satisfiability Testing - {SAT} 2019 - 22nd International
  Conference, {SAT} 2019, Lisbon, Portugal, July 9-12, 2019, Proceedings},
  volume 11628 of {\em Lecture Notes in Computer Science}, pages 250--266.
  Springer, 2019.

\bibitem{Moskewicz2001}
Matthew~W. Moskewicz, Conor~F. Madigan, Ying Zhao, Lintao Zhang, and Sharad
  Malik.
\newblock {Chaff: Engineering an Efficient SAT Solver}.
\newblock In {\em Proceedings of the 38th Annual Design Automation Conference},
  DAC '01, pages 530--535, New York, NY, USA, 2001. ACM.

\bibitem{Nadel2018}
Alexander Nadel and Vadim Ryvchin.
\newblock Chronological backtracking.
\newblock In Olaf Beyersdorff and Christoph~M. Wintersteiger, editors, {\em
  Theory and Applications of Satisfiability Testing - {SAT} 2018 - 21st
  International Conference, {SAT} 2018, Oxford, UK, July 9-12, 2018,
  Proceedings}, volume 10929 of {\em Lecture Notes in Computer Science}, pages
  111--121. Springer, 2018.

\bibitem{Roussel2009}
Olivier Roussel and Vasco~M. Manquinho.
\newblock {Pseudo-Boolean and Cardinality Constraints}.
\newblock In Armin Biere, Marijn Heule, Hans van Maaren, and Toby Walsh,
  editors, {\em Handbook of Satisfiability}, volume 185 of {\em Frontiers in
  Artificial Intelligence and Applications}, pages 695--733. {IOS} Press, 2009.

\bibitem{Sheini2006}
Hossein~M. Sheini and Karem~A. Sakallah.
\newblock {Pueblo: A Hybrid Pseudo-Boolean SAT Solver}.
\newblock {\em {JSAT}}, pages 165--189, 2006.

\end{thebibliography}

\end{document}